# What Robots Do Matters More Than What They Look Like: Task Context Shapes Trust in Educational HRI

Anna-Maria Velentza [1,2] Konstantina Nikou[2] Anne-Gwenn Bosser[1] Nikolaos Fachantidis[2]

***Abstract*—Socially assistive robots (SARs) are increasingly deployed in educational and information-sharing contexts, sup-ported by advances in large language models that enable fluent real-time interaction. Despite the growing diversity of robot embodiments, it remains unclear whether a single robot appearance is appropriate across different interaction tasks or whether trust depends primarily on contextual factors. In this study, we examine how robot appearance and task type jointly influence trust in robots. Using a within-subjects video-based experiment (N = 81), participants evaluated three robots with distinct appearances while performing three educationally relevant tasks: teaching, procedural instruction, and personal-information discussion. Results from repeated-measures analyses show a strong main effect of task on trust, with participants reporting the highest trust during instructional guidance, moderate trust during teaching activities, and significantly lower trust when robots requested personal information. In contrast, robot appearance showed no significant main effect, and the interaction between appearance and task was marginal. These findings suggest that trust in human–robot interaction is shaped more strongly by task context than by physical embodiment alone. By focusing on future educators as end users, this work contributes empirical evidence toward task-aware robot deployment in educational environments and highlights the importance of aligning robot roles and behaviors with interaction goals rather than relying solely on anthropomorphic design.**

[1]Konstantina Nikou nikoukwnstantina@gmail.com and Nikolaos Fachantidis nfachantidis@uom.edu.gr are with the LIRES Robotics Lab, University of Macedonia, GR and School of Educational Social Policies, University of Macedonia, GR. [2]Anna-Maria Velentza annamarakiv@gmail.com and Anne-Gwenn Bosser anne-gwenn.bosser@enib.fr are with the Univ Brest-Bretagne INP, COMMEDIA team, Lab-STICC CNRS UMR 6285, FR.

## I. Introduction

In recent years, there has been increasing research interest in the use of socially assistive robots (SARs) in tasks that involve information sharing, providing instructions, and teaching activities that has been boosted by the application of large language models (LLMs), enriching the production of natural conversation in real time. A variety of different shapes, sizes, color robots are available in the market and in research labs, being demonstrated as capable of fulfilling the above roles. Among those, with which criteria will the educators choose the most appropriate robot for those tasks? Does only one robot design fit all needs?

Trust plays a fundamental role in human-robot interactions (HRI) and thus robots must be considered trustworthy by users to be effectively integrated into daily life [1], [2]. Re-search has shown that robots with a human-like appearance are more likely to be perceived as intelligent, which in turnfosters more positive attitudes and greater social acceptance among humans [3], but not all studies align with the findings that the robot's appearance has an impact on the selective trust [4]. Conversely, other studies [5] suggest that a robot's behavior may play a more significant role than its appearance in shaping social interactions. It is also important to mention that the variety of robots' anthropomorphic designs is wide, and therefore, even robots that are somewhat similar to each other can still be evaluated differently [6].

Despite previous work on robot embodiment and trust in HRI, prior studies often investigate either robot appearance or task context in isolation. As a result, it remains unclear whether trust differences attributed to robot morphology persist when robots perform identical behaviors across multiple interaction contexts. Addressing this gap is important for educational robotics, where the same robot may be expected to perform diverse roles, including instruction, teaching, and socio-emotional interaction.

In this study, we investigate the extent to which robot appearance shapes user trust across interaction contexts, examining whether specific robot embodiments are better suited for teaching, instructional guidance, and personal-information discussions by deploying and comparing three SARs.

Our most important findings are that a) the task strongly influences trust; participants reported the highest trust when robots provided practical instructions, moderate trust during teaching activities, and the lowest trust when asked to disclose personal information. b) Although robot appearance has a limited effect, c) the task outweighs robot type, with the interaction between robot type and task being marginally significant, indicating that trust is shaped more by the nature of the task than by the robot's appearance. These findings, together with another major contribution of the paper, which is the hypothesis testing in the targeted sample of future teachers who will be using robots for educational tasks, highlight the importance of aligning robot behavior and task context to foster trust, rather than relying solely on physical or aesthetic characteristics of the robot.

## II. Related Work

### A. Trust and Social Bond Formation in HRI

Please HRI research has consistently demonstrated that SARs are capable of establishing social bonds with users and supporting trust-based interactions. Prior work shows that the physical embodiment of robots plays an important role in trust formation, contributing to users' sense of comfort and social presence during interaction [7], [8]. Embodiment has also been shown to influence both the quantity and quality of information disclosed to robots, suggesting that users adapt their communicative behavior according to the perceived social characteristics of the agent [9].

Studies with adult populations further demonstrate that users may develop trust relationships with SARs sufficient to disclose sensitive or personal information, even after relatively brief interactions [10]. At the same time, individuals exhibit heterogeneous preferences regarding robot morphology, materials, color, and structural design, as well as robot personality traits, depending on the interaction context [11], [12]. These findings suggest that trust cannot be explained solely by embodiment characteristics but emerges from the interaction between robot design and situational expectations.

### B. Robot Appearance as a Determinant of Perception and Trust

When A substantial body of work has examined how robot appearance shapes user perception. Robot morphology and texture influence first impressions, often conveying implicit cues about intelligence, emotional capacity, and competence be-fore interaction begins [13], [14]. Variations in embodiment, including anthropomorphic, animal-like, or technical designs, have been widely explored as mechanisms for improving acceptance; however, empirical findings remain inconsistent. Some studies demonstrate that appearance interacts with task context when users evaluate robotic services. For example, Choi et al. reported a significant interaction between robot appearance and task type, showing that users evaluated robots differently depending on whether the task involved social service (e.g., serving drinks) or functional work (e.g., vacuuming) [15]. Similarly, appearance-related attributes such as size influence perceived comfort and anxiety, with larger robots increasing interpersonal distance and perceived unease [16], while spatial comfort around robots varies dynamically with movement and proximity [17].

Despite these effects, other research challenges the assumption that appearance alone determines social evaluation. Klu¨ber et al. found that communication modality outweighed appearance and size in shaping robot preference, highlighting interaction capability rather than morphology as the primary determinant of acceptance [18]. Likewise, studies on mind perception show that the actions performed by a robot significantly influence perceived agency, whereas appearance itself may have no measurable effect on perceived agency or experience [5]. Comparable findings emerge in teamwork contexts, where human-likeness and embodiment demonstrated only weak effects on compliance compared to human counterparts [19].

### C. Task Context and Role Expectations in HRI

Recent research increasingly suggests that users evaluate robots relative to task-dependent social expectations. Robot characteristics such as perceived gender have been shown to shift depending on occupational roles, including medical or service-oriented tasks [20]. Such findings imply that users apply human social stereotypes and role schemas when interpreting robotic agents.

Evidence from educational contexts further supports the importance of interaction role. Variations in robot personality and modality influence engagement during teaching activities, even when the same platform is used [12]. Moreover, studies investigating selective trust show that humanoid appearance affects children's trust only in comparative situations; when evaluated independently, appearance does not significantly influence endorsement decisions [21].

### *D. Soft and Socially Expressive Embodiments*

An additional line of work examines robots employing soft materials or pet-like designs intended to promote emotional comfort. Soft, animal-like robots such as Blossom have successfully supported cognitive behavioral therapy exercises with university students, encouraging engagement in wellbeing-related activities [22]. Studies with the therapeutic seal robot PARO report improved mood and social interaction in older adults and people with dementia [23]. Likewise, experimental work with huggable, teddy-like robots demonstrates that reciprocal hugging and soft embodiment increase interaction time and self-disclosure, especially when combined with responsive listening behaviors [24]. Similarly, allowing users to physically customize robots with soft materials has been shown to increase perceived trust toward humanoid platforms such as NAO [25]. Together, these results suggest that the combination of soft, comfort-evoking form factors and attentive social behaviors makes robots particularly effective at eliciting emotional expression and disclosure [14].

## III. Present Study

Taken together, prior literature presents two competing perspectives. On one hand, embodiment and appearance influence comfort, perception, and disclosure behavior. On the other hand, growing evidence suggests that robot behavior, communication, and performed tasks may outweigh physical form in shaping trust and acceptance. Importantly, many studies investigate either appearance or task in isolation, or compare limited embodiment categories within a single interaction context.

Consequently, it remains unclear whether trust in robots is primarily driven by how a robot looks or what a robot is asked to do, particularly in educational scenarios that combine instructional, pedagogical, and socially sensitive interactions.

To address this gap, the present study directly com-pares multiple robot appearances performing identical scripts across distinct task types—teaching, procedural instruction, and personal-information discussion—allowing systematic examination of whether appearance effects persist when task context is controlled.

This study investigates the influence of a) appearance and

b) task. The chosen robots are the Pepper (R1), Nao (R2), and Daisy (R3).

Pepper has been used to deliver how-to instructions and to guide users through procedures (e.g., stepwise demonstrations, tutorial dialogues). Several studies demonstrate Pepper's capability for sequential, spoken guidance, including in rehabilitation and training contexts [26], [27]. However, a study with Peper shows that university students do not have the intention to rely on social robots for learning purposes at the current level of state-of-the-art technology [28]. NAO has been widely used to take the role of a tutor, a therapist, or a peer learner, successfully demonstrating procedures [29], delivering knowledge, and improving levels of enjoyment in university classrooms [30]. Daisy, a flower-shaped design, soft/stuffed-toy style, with "petals" framing a face (tablet screen) and speakers hidden in the petals (See fig.1.c). The robot was designed to assist autistic children with educational or social tasks, e.g., improving social interaction, following instructions, and participating in learning or game activities [31], [32]. Daisy has characteristics similar to fluffy, cute appearance robots such as Paro [23] and Blossom [22], which have been successfully used as companions and mental health supporters across a broad range of ages (see Fig. 1).

Based on the prior literature and the research framework, we formulate the following hypotheses.

**H1.** Robot appearance will influence trust depending on the task type. **H1.a** Participants will exhibit higher trust in humanoid robots (especially NAO) during the teaching activity. **H1.b** Participants will show higher trust in task-oriented robots (especially Peper due to their more authoritative, taller appearance when following step-by-step instructions. **H1.c** Participants will show higher trust in soft robots (Daisy) during personal-information or affective tasks. **H2.** Robots' performed tasks will influence trust.

As previous studies have shown that robots with different characteristics are often perceived differently and may be preferred for different tasks [11], [12].

In more detail, we adapted the three tasks in education activities (teaching, giving instructions, and asking studies stress related personal questions). All participants were exposed in a randomised order to all robots performing all tasks and to measure trust, participants completed post-experiment questionnaires.

While both teaching in terms of lecturing, delivering knowledge and procedural instruction involve interaction with the robot lecturer, the student's perception of the agent's appropriateness may differ. For procedural instruction (T2), students are likely to emphasize the instructor's competence, clarity, and ability to deliver the correct steps. In contrast, for a lecture-style teaching activity (T1), students may place additional weight on subject matter expertise, ability to encourage questions, and support deeper understanding. Prior literature supports this distinction: for example, teaching involves broader facilitation of learning [33], [34] whereas instruction emphasises structured, linear direction. Additionally, trust in lecturers is found to comprise both competence ('can-do') and benevolence ('will-do') dimensions [35].

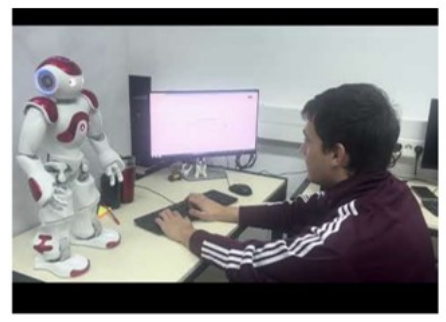
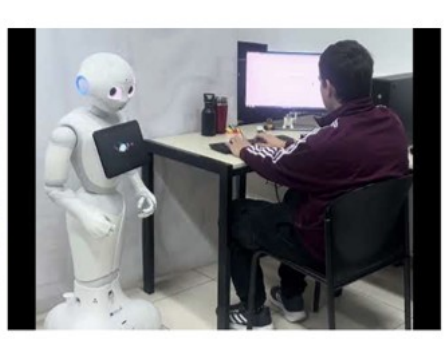
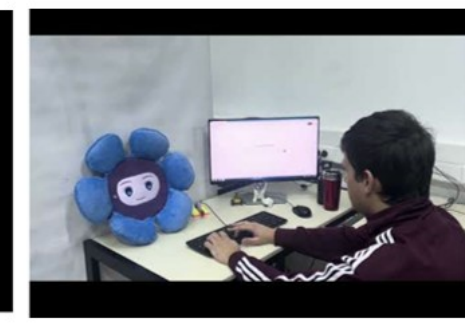

Fig. 1. Different Robots same Taks: Example of the Instruction's Task

## IV. Method

### A. Participants

N = 81 (82% women and 18% men) undergraduate students in the School of Educational and Social Policies aged 18–28 studying to become educators, with proficiency in the Greek language that the experiment was conducted, anonymously and voluntarily participated after agreeing to the approved university's ethics committee guidelines.

### B. Design

The study was administered through the university's secure online learning platform, which students regularly use for coursework. Using a familiar system minimized usability-related confounds and reduced variability due to platform novelty. All data collection occurred within this protected environment. Prior to the online phase, a co-author who was the course instructor presented the three robots, Pepper, NAO, and Daisy, during a scheduled class session. This ensured that all students had direct physical exposure to the robots before the experiment, thereby reducing novelty effects and unequal familiarity across conditions. The instructor introduced the study only after the teaching session had concluded, emphasizing that participation was voluntary and unrelated to course assessment in order to mitigate perceived coercion or demand characteristics. A link to the study was projected at the end of class and remained active for 15 days. Students accessed a dedicated course-platform page containing detailed study information and ethics documentation before providing informed consent. Participation occurred asynchronously and outside class time.

We employed a 3 × 3 fully within-subjects design with robot embodiment (Pepper-R1, NAO-R2, Daisy-R3) and interaction task (teaching-T1, instructional guidance-T2, per-sonal conversation-T3). Each robot delivered identical scripts across the three themes, yielding nine video stimuli, following the methodological paradigm of [36], who measured user preference through different human-robot interactions. Script content was held constant to isolate the effect of embodiment while controlling for semantic variation. Presentation order was fully randomized per participant across both robot and theme to mitigate order, fatigue, and sequence effects. To control for vocal confounds, speech output across robots was calibrated to equivalent pitch and rate. Calibration was validated in a pilot study (N = 10) with independent lab members to ensure perceptual comparability and prevent voice-related bias.

### C. Proceedure

After agreeing to the informed consent, participants viewed all nine videos in randomized order. The video-based methodology was chosen to maintain strict control over robot behavior, dialogue content, and timing across platforms. This approach ensured that any differences in participant evaluations could be attributed to robot embodiment or task context rather than uncontrolled behavioral variation. To ensure standardized stimulus exposure and prevent selective skipping, videos had to be viewed in full before progression was enabled. The same questionnaire sequence followed each video, and all questionnaire items were mandatory before proceeding. Demographic questions appeared at the end. No time limit was imposed, reducing time-pressure effects and supporting response quality. The forced-completion structure prevented missing data within conditions and ensured balanced observations across all factors. The within-subjects de-sign minimized between-participant variability and increased statistical power [37] while controlling for individual differences in prior robot experience, attitudes toward technology, and communication preferences [38].

**Task Scripts**

*Task 1: Teaching Activity (T1)*

The robot introduces the concept of Artificial Intelligence (AI) as the ability of machines to perform tasks that require human-like intelligence, such as answering questions, analyzing data, learning from information, and recognizing pat-terns. It explains examples of AI applications in education, healthcare, and industry, and asks the participant if they have any questions.

**Robot:** "Hello! Today we will talk about Artificial Intelligence, also known as AI. AI is the ability of machines, like me, to perform tasks that require human intelligence... It is used to improve services in education, health, and industry. Do you have any questions so far?"

**Human:** "No!"

[Lasted 0.58min]

*Task 2: Instruction Activity (T2)*

The robot provides step-by-step instructions on how to use ChatGPT. It guides the participant to open the app or website, locate the input screen, and type a sample question. The participant follows along as the robot explains how ChatGPT can respond to different types of questions and provide practical advice.

**Robot**: "Hello! Today, we will show you how to use Chat-GPT. Start by opening the app or website. Do you see the input screen? There, you can type your question... Try asking something simple, like 'How can I organize my time better?'... Great! ChatGPT will give you an answer. You can ask anything—from study ideas to daily advice."

[Lasted 1.02min]

*Task 3: Personal Questions Activity (T3)*

The robot engages in a short personal conversation about the participant's studies and emotions. It asks whether they feel stressed about their courses, discusses what situations cause stress, and explores coping strategies. The robot responds empathetically and validates the participant's feelings.

**Robot:** "Hello! I'd like to ask you a few questions about your studies. Do you feel anxious about your classes?"

**Human:** "Yes, sometimes I do."

**Robot**: "That's completely normal, especially when working on something important. Which courses or situations make you feel most stressed?"

**Human**: "Probably those with a lot of assignments or exams."

**Robot**: "That's understandable. Have you tried any strategies to manage your stress?"

**Human**: "I try to relax by watching a movie or talking to a friend."

**Robot**: "That sounds very effective! Thank you for sharing that with me."

[Lasted 1.04min]

*Data Collection*

The same 10 questionnaire items were presented after each video interaction and rated on a five-point Likert scale. A multidisciplinary group of psychologists, engineers, and teachers developed the pre- and post-questionnaires by combining questions (items) from three established questionnaires, as follow:

1) **I would like the robot to give me usage instructions**.

(*maps to instruction-intent / willingness items; [39]*)

2) **I consider the robot capable of giving me usage instructions.**

(*competence/performance item; [40]*)

3) **I trust the robot to give me correct usage instructions.**

(*explicit trust/performance item; [39]*)

4) **I believe the robot can give usage instructions better than a human.**

(*relative-performance / comparative competence item; [41]*

5) **I believe the robot can successfully provide usage instructions.**

(*success / reliability phrasing; [40]*)

6) **I feel the robot can communicate usage instructions clearly**.

(*communication clarity; related to perceived intelligibility / communication items in Godspeed [42]*

7) **I believe the robot can communicate effectively with me.**

(*general communication effectiveness; maps to per-ceived intelligence/communication constructs [42]).*

8) **I believe the robot can provide appropriate and understandable usage instructions to people**. (*appropriateness & comprehensibility; performance + usability wording [42], [40])*

The elements were adapted to the Greek language following the technique of bilingual speakers. Content validity was assessed using Lawshe's Content Validity Ratio (CVR). method, in which subject matter experts evaluate the essentiality of each item. Five subject matter experts from the LIRES Lab independently reviewed the questionnaire items. The resulting CVR met the critical threshold for five experts (CVR = .883, two-tailed p = .05), indicating acceptable content validity according to the recalculated critical values reported by James Wilson and colleagues in their methodological update of Lawshe's framework [43]. Data were analyzed using Jamovi software (version 2.321). Using

### D. Data Analysis

Trust scores were computed by averaging item responses for each condition. Descriptive statistics (means and standard deviations) were first examined to identify patterns across robots and tasks. Given the within-subjects design with nine repeated conditions, the sample size provides adequate statistical power to detect medium effects in repeated-measures analyses, while controlling for inter-individual variability.

To examine the effects of embodiment and interaction context on trust, a two-way repeated-measures ANOVA was conducted with ROBOT (3 levels) and TASK (3 levels) as within-subject factors. Assumptions of sphericity were evaluated using Mauchly's test. When violations occurred, Greenhouse–Geisser and Huynh–Feldt corrections were ap-plied to adjust the degrees of freedom. Effect sizes are reported using partial eta squared ($\eta^2_\rho$). Post hoc pairwise comparisons were conducted with Bonferroni adjustments to control for multiple comparisons. Additionally, Pearson correlation analyses were performed on mean trust scores across all robot–task combinations to examine the consistency of participants' trust evaluations across contexts.

## V. RESULTS

Cronbach's α coefficients were computed for each robot–task combination to assess internal reliability. Values ranged from .85 to .95, indicating excellent internal consistency across all conditions. Specifically, reliability co-efficient were α = .924 for R1T1, α = .910 for R1T2, α = .937 for R1T3, α = .853 for R2T1, α = .945 for R2T2, α = .948 for R2T3, α = .913 for R3T1, α = .935 for R3T2, and α = .925 for R3T3 (see Fig. 2).

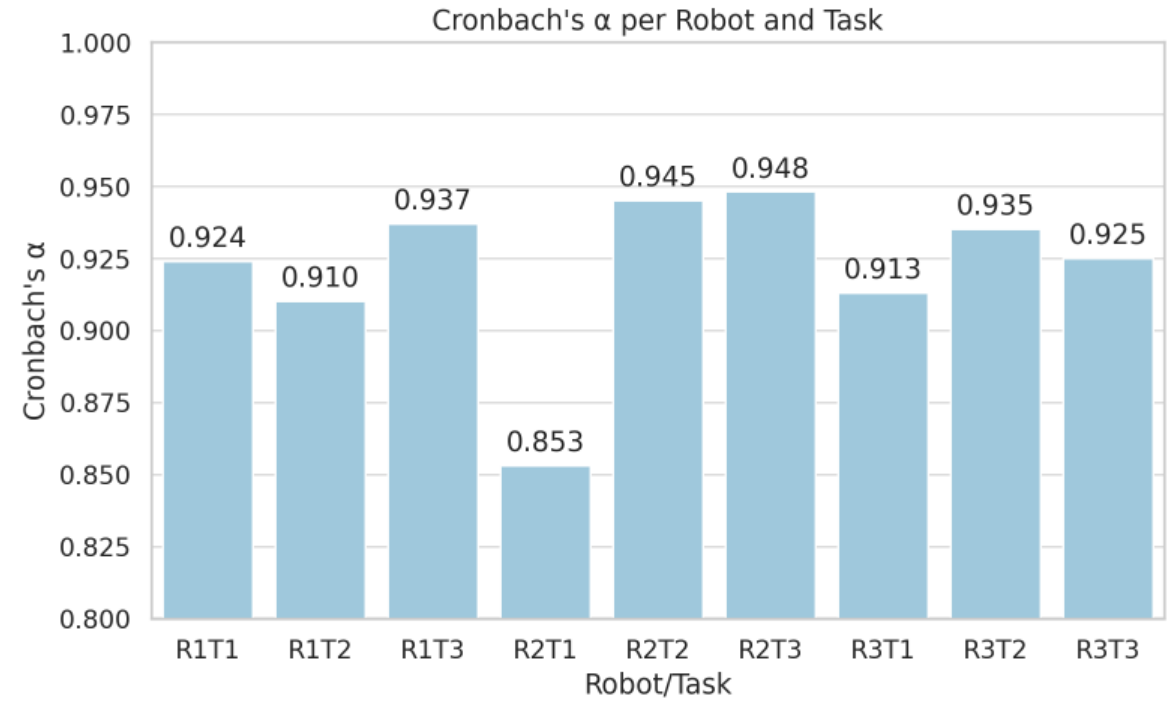

Fig. 2. Cronbach's α per Robot and Task

### *Descriptive Statistics*

Descriptive analyses of the mean trust scores across robots and tasks revealed clear patterns (see Fig. 3). During the Teaching activity (T1), trust ratings were relatively similar across robots (R2: M = 3.21, SD = 0.60; R1: M = 3.11, SD = 0.74; R3: M = 3.10, SD = 0.71). In the Instructions activity (T2), trust increased for all robots (R1: M = 3.53, SD = 0.55; R2: M = 3.41, SD = 0.73; R3: M = 3.35, SD = 0.67). In contrast, during the Personal Information activity (T3), trust ratings declined (R2: M = 2.91, SD = 0.87; R1: M = 2.88, SD = 0.81; R3: M = 2.86, SD = 0.78).

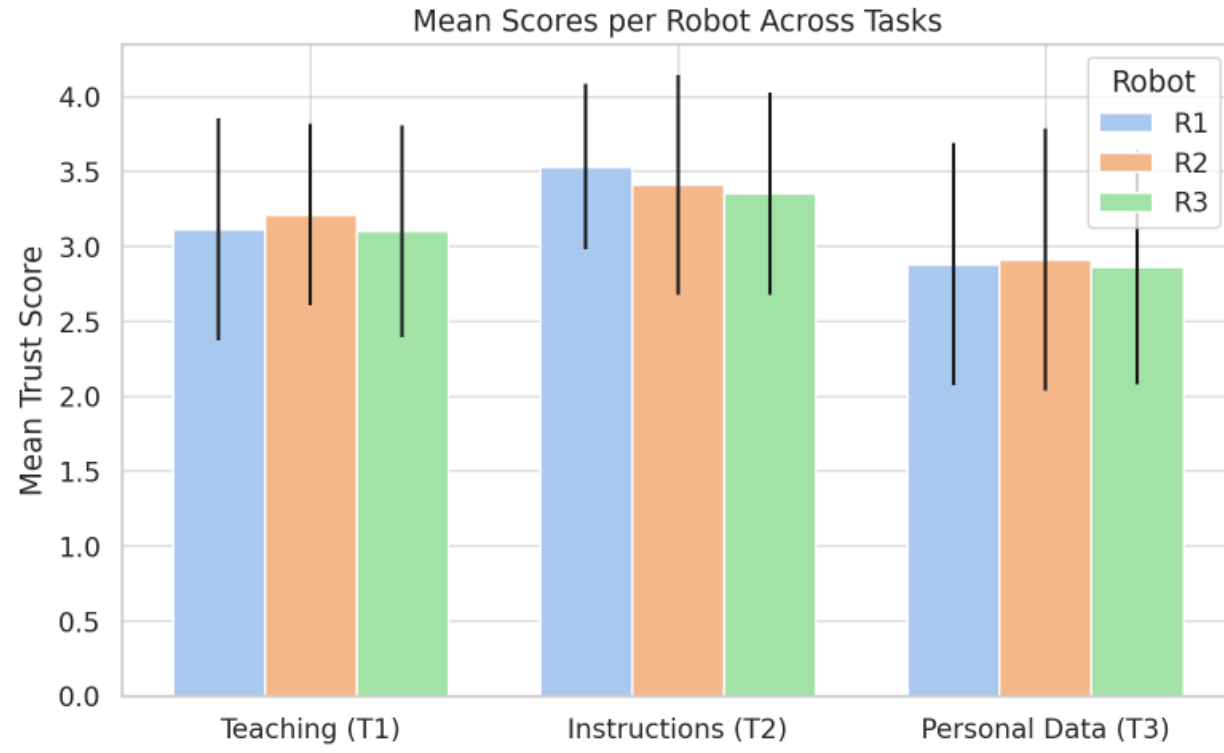

Fig. 3. Mean Scores per Robot Across Tasks

These findings indicate that participants reported the highest levels of trust when robots provided practical instructions, moderate trust during teaching, and the lowest trust when robots requested personal information.

A repeated-measures ANOVA was conducted with ROBOT (R1, R2, R3) and TASK (T1, T2, T3) as within-subjects factors. Results revealed a significant main effect of TASK, $F$ (1.88, 148.18) = 61.88, $p < .001$, $\eta^2 = .439$, indicating that trust levels differed significantly across activities. The main effect of ROBOT was not significant, $F$ (1.89, 156.75) = 3.82, $p = .024$, $\eta 2 = .046$. The ROBOT x TASK interaction was marginal $F$ (2.94, 232.64) $p = .034$, $\eta^2 = .036$, suggesting a limited combined influence of robot type and task context on trust ratings (see Fig. 4).

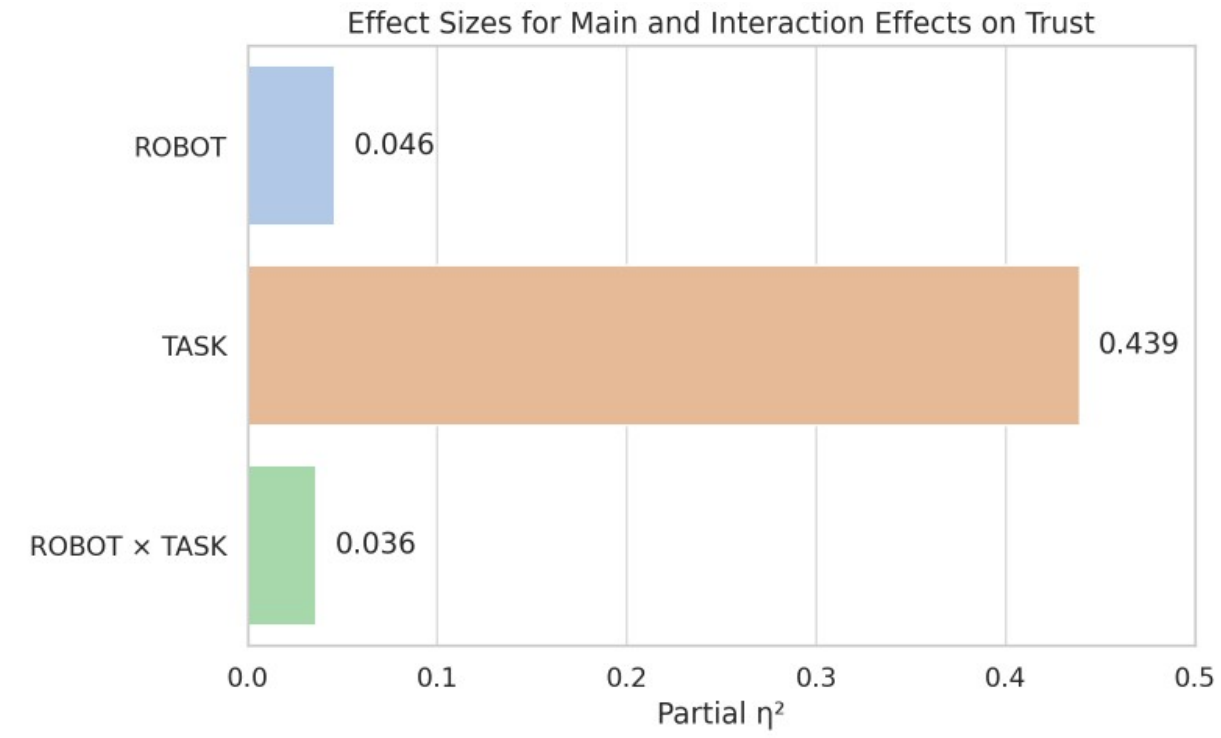

Fig. 4. Effect Sizes (Partial η2 ) for Main and Interaction Effects

Mauchly's test indicated that the assumption of sphericity was met for ROBOT ($p = .731$), marginally violated for TASK ($p = .069$), and violated for the ROBOT × TASK interaction ($p<.001$). Accordingly, Greenhouse–Geisser and Huynh–Feldt corrections were applied to adjust degrees of freedom where appropriate.

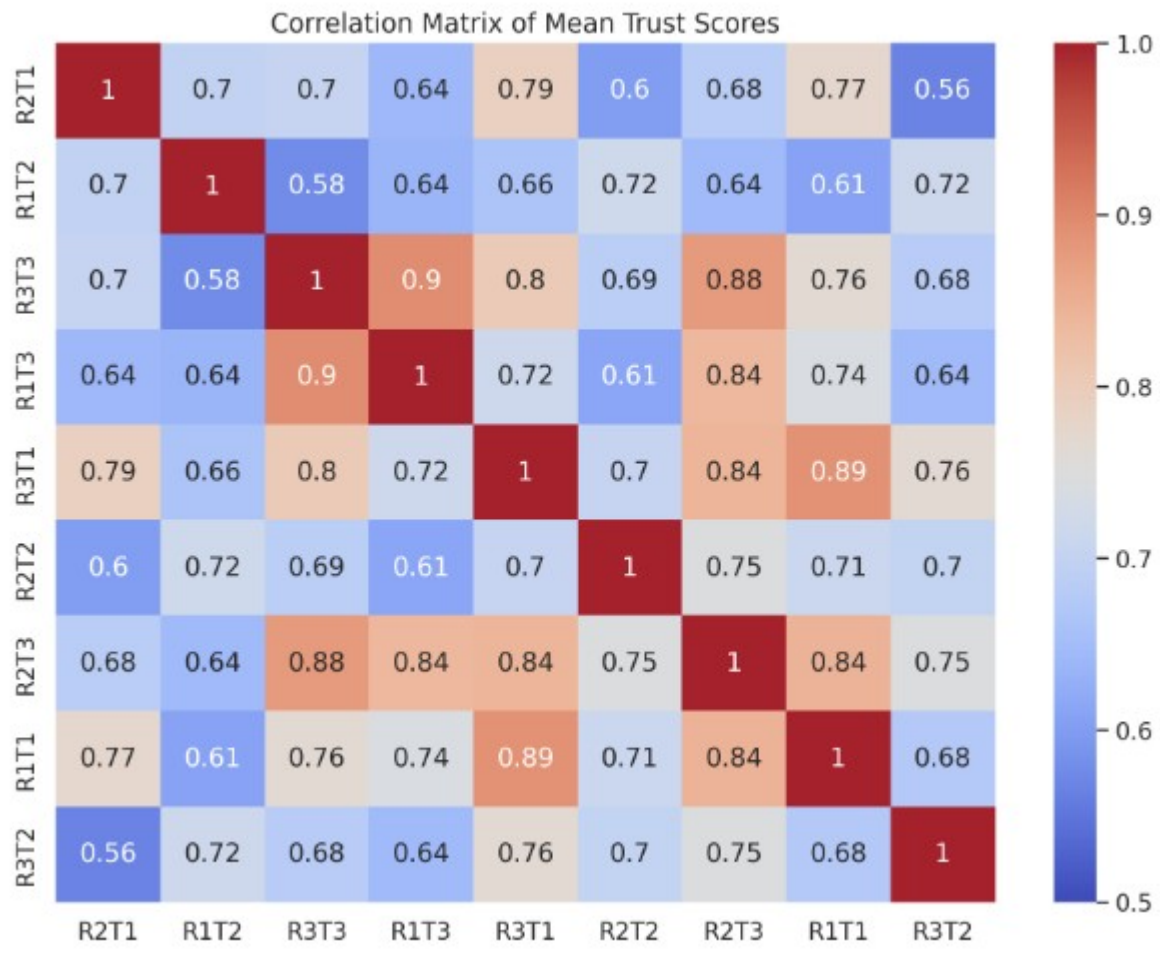

Fig. 5. Correlation Matrix of Mean trust Scores

*Post Hoc Comparisons*

Pairwise comparisons between robots revealed no statistically significant differences in trust. Differences between R1–R2 ($p<.05$), R1–R3 ($p = .069$), and R2–R3 ($p = .062$) did not reach significance after Bonferroni adjustment. Thus, the type of robot did not meaningfully influence participants' trust.

Post hoc comparisons for TASK (see Fig. 6) showed that T2 (Instructions) produced significantly higher trust scores than both T1 (Teaching) and T3 (Personal Information), all $p < .001$. Additionally, T1 elicited higher trust than T3, $p < .001$. These results confirm that task type exerted a strong influence on trust ratings.

A correlation matrix of mean trust scores (Fig. 5) revealed strong positive associations across all robot–task combinations, with $r = .56–.89$ and $p < .001$. Participants who accepted a given robot in one context tended to accept the same and other robots across different contexts, indicating consistent trust perceptions.

Estimated marginal means confirmed that overall trust did not differ significantly between robots (R1: $M = 3.17$, R2: $M = 3.17$, R3: $M = 3.10$). However, trust varied clearly across tasks: highest for Instructions (T2, $M = 3.43$), followed by Teaching (T1, $M = 3.14$), and lowest for Personal Information (T3, $M = 2.88$). As shown in Fig. 6, this pattern was consistent across all robots (Fig.7), suggesting that task type outweighed robot type in shaping trust levels.

Overall, all measures showed excellent reliability ($\alpha \geq .85$). Task type was the strongest determinant of trust, explaining substantial variance ($\eta^2_\rho = .439$). Robot type had no significant main effect, and the ROBOT × TASK interaction was marginal. Participants exhibited greater trust when robots provided practical instructions, moderate trust when robots provided practical instructions, moderate trust in teaching contexts, and lower trust when personal data were requested.

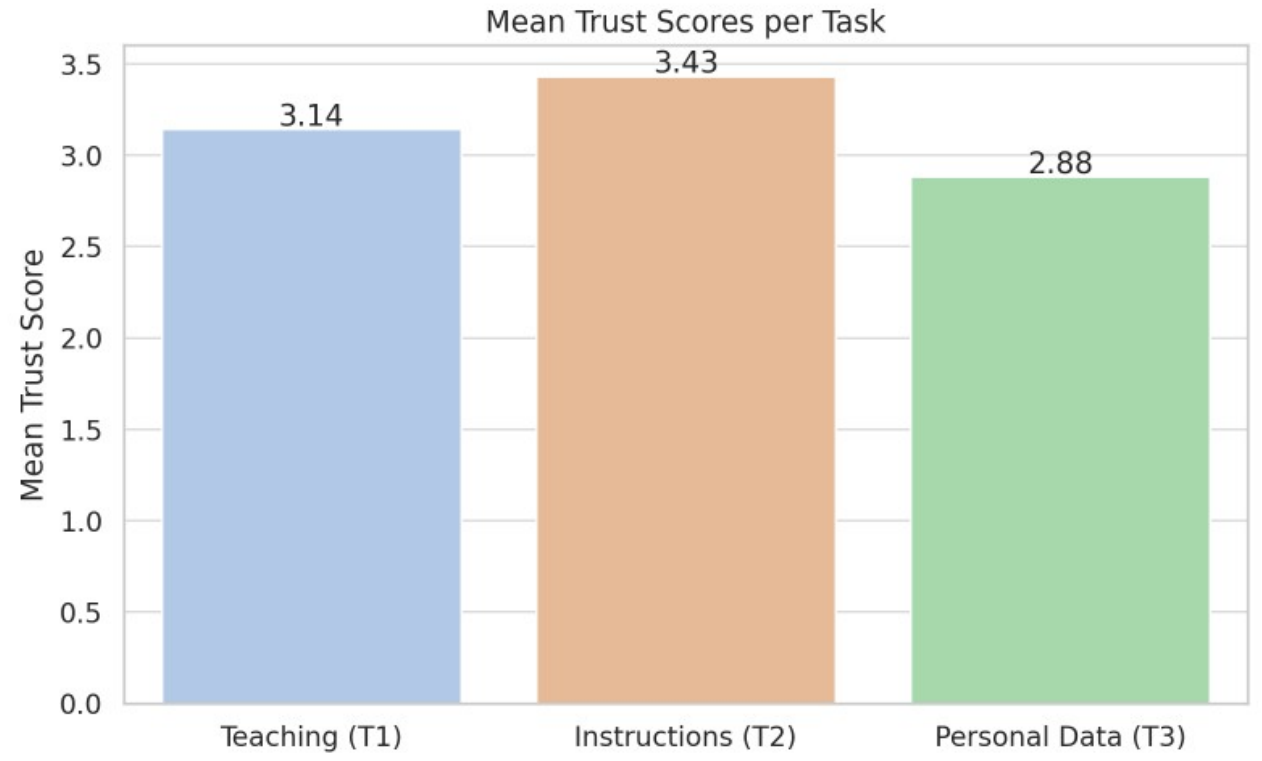

Fig. 6. Post Hoc Comparisons: Mean differences between tasks (T 2 > T 1 > T 3)

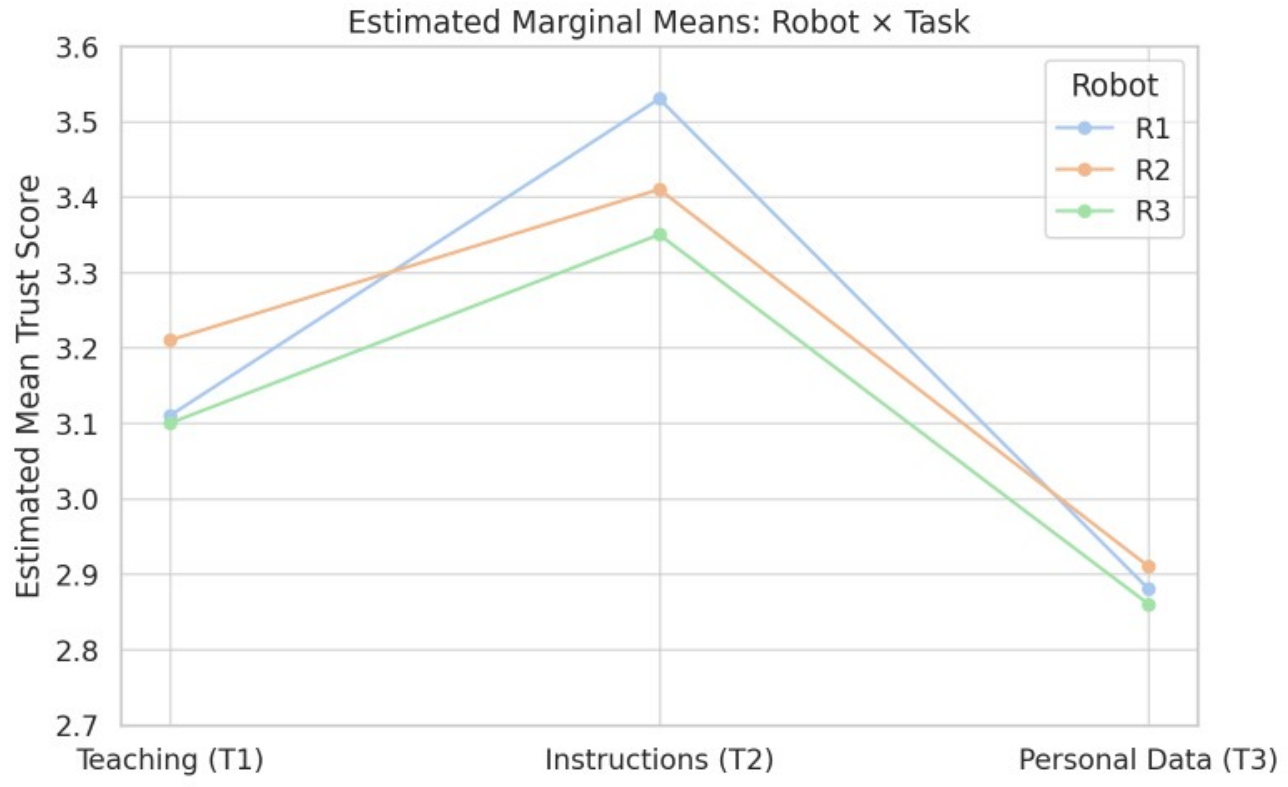

Fig. 7. Interaction trust scores per task for each robot

## VI. DISCUSSION AND CONCLUSIONS

This study examined whether robot appearance or interaction context plays a stronger role in shaping users' trust toward socially assistive robots. Contrary to common assumptions in HRI that anthropomorphic or socially expressive designs inherently foster trust, our findings demonstrate that task context is the primary determinant of trust, while robot appearance exerts only a limited influence.

Across all three robots, participants consistently reported the highest trust when robots delivered procedural instructions, followed by teaching activities, and the lowest trust when robots engaged in personal-information discussions. This pattern suggests that users calibrate trust according to additional factors, such as potentially the perceived task risk and expectations of competence, rather than physical embodiment alone. Those results are consistent with the findings of [44], suggesting that contextual factors trigger a stronger influence even than biases arising from appearance-related characteristics. This interpretation aligns with trust calibration theories in automation research, which propose that users adjust their reliance on artificial agents according to perceived task risk, uncertainty, and expected system competence rather than agent appearance alone. Instructional tasks are structured, goal-oriented, and objectively verifiable, which may reduce uncertainty and promote confidence in robotic performance. In contrast, personal conversations involving emotional disclosure introduce higher social and privacy sensitivity, leading participants to apply stricter trust criteria.

The intercorrelations among mean trust scores were uniformly positive and statistically significant ($p < .001$), indicating a coherent perception pattern across conditions. The hypotheses related to matching robot appearance to task based on each robot's external characteristics were not confirmed. Participants who reported higher trust in a robot during one task were likely to express similar levels of trust toward the same robot in other tasks. This finding implies that trust in social robots may represent a stable, generalized attitude rather than a purely task-dependent evaluation.

The absence of a significant main effect of robot appearance aligns with prior work suggesting that robot behavior and performed actions outweigh morphology in shaping social perception. Even though humanoid robots are often assumed to enhance credibility or intelligence perceptions, our results indicate that such advantages may diminish when identical behaviors and scripts are controlled across platforms. The marginal interaction effect further suggests that appearance may slightly modulate expectations but does not fundamentally determine trust formation.

Importantly, these findings contribute to an ongoing debate in HRI regarding "one-robot-fits-all" deployment strategies. Rather than selecting robots solely based on anthropomorphism or aesthetic appeal, designers and educators may benefit more from aligning robot capabilities and interaction roles with task demands. For example, robots may be readily accepted as instructional guides but require additional behavioral or relational mechanisms before being trusted in emotionally sensitive interactions.

Another key contribution of this work lies in examining future teachers as a target population. As educators represent likely adopters of classroom robots, understanding how this group evaluates trust across pedagogical tasks provides eco-logically relevant insights for educational robotics deployment. The consistency of trust patterns across robots suggests that acceptance barriers may emerge less from embodiment choices and more from how robots are positioned within educational workflows. It is also important to note that all three robots were introduced to the students by their instructor, and thus it is reasonable to infer that this increased their perceived trust in them, while by introducing them, we aimed to avoid the novelty effect. In the future, it would be interesting to test robots with similar characteristics that future teachers might not be familiar with.

The results suggest that designers of educational robots should prioritize aligning robot behavior and interaction capabilities with task requirements rather than focusing primarily on anthropomorphic appearance. Robots used for procedural guidance or structured instruction may be accepted regardless of embodiment, while tasks involving emotional disclosure or personal information may require additional design considerations such as transparency, privacy assurances, or enhanced socio-emotional behaviors.

*Limitations*. The study employed a video-based methodology, enabling strict experimental control. Robot behaviors were intentionally script-controlled and equivalent across platforms to isolate appearance effects. While methodologically necessary, this likely reduced natural differences in expressiveness, autonomy, and interaction style that may influence trust in real deployments, since, according to [45] video-mediation can also mitigate the effect of the robot's appearance. In addition, the participant sample consisted of undergraduate students from educational programs. Although appropriate for examining future professional users, this relatively homogeneous group limits generalization to other populations, such as children, experienced teachers, or clinical users. Moreover, although the sample appears to be gender imbalanced, it is representative of future teachers in Greek universities.

*Future Work*. We encourage future research to examine whether the present findings generalize to real-time interactions and longitudinal exposure. Trust in HRI is dynamic and evolves through repeated encounters; therefore, long-term studies involving sustained classroom interaction may reveal embodiment effects that do not emerge in controlled short-term evaluations. Additionally, investigating a broader range of robot embodiments, including mobile service robots, minimalistic agents, and highly anthropomorphic platforms, would further clarify the relationship between morphology and task suitability. Finally, studies involving different user populations, such as experienced teachers, younger students, or clinical users, would help determine whether trust patterns observed in future educators generalize to other stakeholder groups.

## Acknowledgment

Discobot project, supported by the Region Bretagne and the Conseil departemental du Finistere.